\documentclass{article} 
\usepackage{iclr2015,times}
\usepackage{hyperref}
\usepackage{url}
\usepackage{amsmath}
\usepackage{amssymb}
\usepackage[nolist]{acronym} 
\usepackage{cancel}
\usepackage{graphicx}
\usepackage{todonotes}
\usepackage{tikz}
\usetikzlibrary{positioning,shapes}

\title{Learning Stochastic Recurrent Networks}

\author{
    Justin Bayer  \\
    Lehrstuhl f\"ur Echtzeitsysteme und Robotik \\
    Fakult\"at f\"ur Informatik\\
    Technische Universit\"at M\"unchen \\
    \texttt{bayer.justin@googlemail.com}\\
\AND
    Christian Osendorfer \\
    Institut f\"ur Regelungstechnik \\
    Leibniz Universit\"at Hannover \\
\texttt{christian.osendorfer@rt.uni-hannover.de}
}

\iclrconference 

\newcommand{\eq}[1]{\begin{align*}#1\end{align*}}
\newcommand\numberthis{\addtocounter{equation}{1}\tag{\theequation}}

\newcommand{\bex}[1]{\begin{framed}\begin{sc}Example: #1\end{sc}\\}
\newcommand{\eex}{\end{framed}}

\newcommand{\bi}{\begin{itemize}}
\newcommand{\ei}{\end{itemize}}

\DeclareMathOperator*{\argmax}{argmax}

\newcommand{\bh}{\mathbf{h}}

\newcommand{\bx}{\mathbf{x}}
\newcommand{\by}{\mathbf{y}}
\newcommand{\bz}{\mathbf{z}}

\newcommand{\lT}{_{1:T}}

\newcommand{\ltt}{_{1:t}}
\newcommand{\Win}[0]{\mathbf{W}_{\text{in}}}
\newcommand{\Wout}[0]{\mathbf{W}_{\text{out}}}
\newcommand{\Wrec}[0]{\mathbf{W}_{\text{rec}}}
\newcommand{\bout}[0]{\mathbf{b}_{\text{out}}}
\newcommand{\bhid}[0]{\mathbf{b}_{\text{hidden}}}

\newcommand{\Lnll}[0]{\mathcal{L}_{\text{NLL}}}

\newcommand{\Expc}[2]{\mathbb{E}_{#1}[#2]}

{}
\newcommand{\belowspace}{}{}

\begin{document}

\maketitle

\begin{abstract}
Leveraging advances in variational inference, we propose to enhance recurrent neural networks with latent variables, resulting in \acp{STORN}.
The model i) can be trained with stochastic gradient methods, ii) allows structured and multi-modal conditionals at each time step, iii) features a reliable estimator of the marginal likelihood and iv) is a generalisation of deterministic recurrent neural networks.
We evaluate the method on four polyphonic musical data sets and motion capture data.

\end{abstract}

\section{Introduction}
\label{sec:intro}

\acp{RNN} are flexible and powerful tools for modeling sequences.
While only bearing marginal existence in the 1990's, recent successes in real world applications~\citep{graves2013generating, graves2013speech, sutskever2014sequence, graves2008unconstrained, cho2014learning} have resurged interest.
This is partially due to architectural enhancements~\citep{hochreiter1997long}, new optimisation findings~\citep{martens2011learning,sutskever2013importance,bengio2012advances} and the increased computional power available to researchers.
\acp{RNN} can be employed for a wide range of tasks as they inherit their flexibility from plain neural networks.
This includes universal approximation capabilities, since \acp{RNN} are capable of approximating any measureable sequence to sequence mapping and have been shown to be Turing complete~\citep{hammer2000approximation,siegelmann1991turing}.

One typical application is to let an \ac{RNN} model a probability distribution over sequences, i.e. $p(\bx\lT)$.
This is done by writing the distribution in cascade form,
\eq{
    p(\bx\lT) & = \prod_{t=0}^{T-1} p(x_{t+1}|\bx\ltt),
}
where $\bx_{1:0} = \emptyset$.
Each $p(x_{t+1}|\bx\ltt)$ is then represented by the output of an \ac{RNN} at a single time step, identifying each of its components with the statistics of the distribution.
A simple example is that of a Bernoulli, i.e.
\eq{
    p(x_{t+1, k} = 1|\bx\ltt) &= \eta_k(\bx\ltt) \numberthis \label{eq:bernoulli}
}
where $x_{t+1, k}$ corresponds to the $k$'th component of the $t+1$'th time step of $\bx$ with $k=1,\dots,\omega$ and $t=1,\dots,T$.
Each $\eta_k(\bx\ltt)$ is the $k$'th output of some \ac{RNN} at time step $t$, constrained to lie in the interval $(0, 1)$.
Learning such an \ac{RNN} then boils down to minimising the negative log-likelihood of the data with respect to the parameters of the network.

This framework gives practitioners a powerful tool to model rich probability distributions over sequences.
A common simplification is a na\"ive Bayes assumption that the individual components factorise:
\eq{
    p(x_{t+1}|\bx_{1:t}) &= \prod_k p(x_{t+1, k}|\bx_{1:t}).
}
While sufficient for many applications, reintroduction of dependency among the components of $x_t$ leaves room for improvement.
This is especially true for sequences over spaces which are high dimensional and tightly coupled.
The approach taken by~\cite{graves2013generating} is to use a mixture distribution for $p(x_t|\bx_{1:t-1})$.
Arguably powerful enough to model any dependency between the components of $x_t$, a drawback is that the number of parameters scales at least linearly with the number of chosen mixture components.

Models based on restricted Boltzmann machines and variations~\citep{boulanger2012modeling,boulanger2013high,sutskever2008recurrent} provide a solution to this as well, yet come with tighter restrictions on the assumptions that can be made.
E.g. RBMs are restricted to model data using posteriors from the exponential family~\citep{welling2004exponential}, make use of an intractable objective function and require costly MCMC steps for learning and sampling.

In this work, we propose to consider adding latent variables similar to \cite{tang2013learning} to the network.
Using \ac{SGVB}~\citep{rezende2014stochastic,kingma2013auto} as an estimator, we train \acp{RNN} to model high dimensional sequences.

\section{Preliminaries}
In this section we will recap the basis of our method.
We will first describe the used model family, that of recurrent neural networks and then the estimator, stochastic gradient variational Bayes (SGVB).
\subsection{Recurrent Neural Networks}
\label{sub-sec:rnn}

Given an input sequence $\mathbf{x} = (x_1, \dots, x_T), x_t \in \mathbb{R}^\kappa$ we compute the output sequence of a \ac{sRNN} $\mathbf{y} = (y_1, \dots, y_T), y_t \in \mathbb{R}^\omega$ via an intermediary hidden state layer $\mathbf{h} = (h_1, \dots, h_T), h_t \in \mathbb{R}^\gamma$ by recursive evaluation of the following equations:
\eq{
    h_t &= f_h(x_t \Win + h_{t-1} \Wrec + \bhid), \numberthis \label{eq:hidden}, \\
    y_t &= f_y(h_t \Wout + \bout). \numberthis \label{eq:output}
}
The set of adaptable parameters is given by $\theta = \{\Win, \Wrec, \Wout, \bhid, \bout \}$.
$f_h$ and $f_y$ are transfer functions introducing nonlinearity into the computation.

Adaptation of the network's behaviour can be done by optimising a loss function with respect to the network's parameters with gradient-based schemes.
Consider a data set of finite size, i.e. $\mathcal{D} = \{(\bx^{(i)}\lT)\}_{i=1}^{I}$ on which the loss operates.
In a setting as in Equation~(\ref{eq:bernoulli}) a reasonable choice is the negative log-likelihood given by $\Lnll(\theta) = -\sum_{i=1}^I \sum_{t=1}^T \log p(x_t|\bx_{1:t-1})$.

\subsection{Stochastic Gradient Variational Bayes}
\label{sec:sgvb}

\ac{SGVB} was introduced independently by \cite{rezende2014stochastic} and \cite{kingma2013auto}.
For this paper, we will review the method briefly in order to introduce notation.
We are interested in modelling the data distribution $p(\bx)$ with the help of unobserved latent variable $\bz$ represented as a directed graphical model, i.e. $p(\bx) = \int p(\bx|\bz)p(\bz)d\bz$.
The integral is in general intractable, which is why we will use a variational upper bound on the negative log-likelihood for learning.
\eq{
    -\log p(\bx) &= -\log \int p(\bx|\bz)p(\bz)dz \\
    &= -\log \int {q(\bz|\bx) \over q(\bz|\bx)} p(\bx|\bz)p(\bz) dz \\ 
    &\le KL(q(\bz|\bx)||p(\bz)) - \Expc{z \sim q(\bz|\bx)}{\log p(\bx|\bz)} =: \mathcal{L}.
}
where $KL(q||p)$ denotes the Kullback-Leibler divergence of $p$ from $q$.
In this case, we call $q$ the \emph{recognition model} since it allows for fast approximate inference of the latent variables $\bz$  given the observed variables $\bx$.
Note that $q$ is a variational approximation of $p(\bz|\bx)$, which is the inverse of the \emph{generating model}\footnote{We use the non standard term ``generating model'' for $p(\bx|\bz)$ to distinguish it more clearly from the generative model $p(\bx)$.}
$p(\bx|\bz)$ that cannot be found in general.

Both the recognition and the generating model can be chosen arbitrarily in their computational form with the possibility to represent probability distributions as outputs and stochastic training being the only requirements.
In order to minimise the upper bound of the negative log-likelihood $\mathcal{L}$ with numerical means, it is convenient to choose parametric models.
In that case we write $p(\bx|\bz, \theta^g)$ and $q(\bz|\bx, \theta^r)$ to make the dependency on the respective parameter sets explicit.
Learning good parameters can then be done by performing stochastic optimization of $\mathcal{L}$ with respect to both $\theta^r$ and $\theta^g$, where the expectation term is approximated by single draws from $q$ in each training step.

Designing a model is then done by the following steps: (1) Choice of a prior $p(\bz)$ over the latent variables. (2) Choice of a recognition model $q(\bz|\bx, \theta^r)$.
The Kullback-Leibler divergence between the prior and the recognition model has to be tractable and efficient to compute. (3) Choice of a generating model $p(\bx|\bz, \theta^g)$, which is often given by the type of data under investigation.

An important question is that of the representation capabilities of such a model.
It turns out that if the distribution $p(x|z)$ is a universal \emph{function} approximator, so is the overall model.
An argument for the one-dimensional case is as follows.
Assume random variables $x$ and $z$ with respective distribution functions $F_x$ and $F_z$.
According to the inverse transform technique theorem \citep{grimmett1992probability}, $u = F^{-1}_x(x)$ will be uniformly distributed over the interval $[0, 1]$ and so will be $u' = F^{-1}_z(z)$.
Equating gives $F^{-1}_z(z) = F^{-1}_x(x) \Rightarrow F_x(F^{-1}_z(z)) = x$.
Therefore setting $p(x|z) := \delta(x = f(z))$ with $F = F_x \circ F^{-1}_z$ makes $p(x) = 
\int_z p(x|z)p(z)dz$.
An extension to the multidimensional case can be done by applying the above to the individual factors of a cascade decomposition and requiring $x$ and $z$ to be of the same dimensionality.
The burden is then on the learning algorithm to find a good approximation for $F$.

\section{Methods}

We propose to combine \ac{SGVB} and \acp{RNN} by making use of an \ac{sRNN} for both the recognition model $q(z_t|\bx_{1:t-1})$ and the generating model $p(x_t|\bz_{1:t})$.

\subsection{The Generating Model}
More specifically, the generating model is an \ac{sRNN} where the latent variables form additional inputs:
\eq{
    h_t &= f_h(x_t \Win^g + z_t \Win'^g + h_{t-1} \Wrec + \bhid) \numberthis \label{eq:hidden_storn} \\
}
which replaces Eq.~(\ref{eq:hidden}).
We let $y_t$ from Eq.~(\ref{eq:output}) represent the necessary statistics to fully determine $p(x_{t+1}|\bx\ltt)$.

Note that the model reduces to an \ac{sRNN} as soon as we remove any latent variables, e.g. by setting $\Win'^g = \mathbf{0}$.
Hence, such a model generalises \acp{sRNN}.

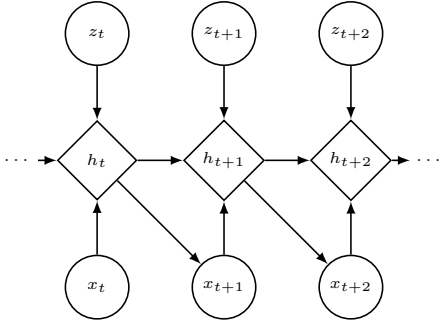
\begin{figure}
\label{fig:graphmodel}
\centering
\tiny
\begin{tikzpicture}[
    -latex,
    auto,
    node distance=12em,
    on grid,
    semithick]

\tikzstyle{deterministic} = [diamond, draw=black, fill=white, text=black, minimum width=5em, minimum height=5em, line width=.1em]
\tikzstyle{stochastic} = [circle, line width=.1em, draw=black, fill=white, text=black, minimum width=3.5em, minimum height=4em]
\tikzstyle{invisible} = [draw=none, fill=none, text=black]
\tikzstyle{stochastic2} = [circle, line width=.1em, draw=black, top color=white, bottom color=red, text=black, minimum width=4em, minimum height=4em]

\node[stochastic] (Z1) {$z_t$};
\node[stochastic] (Z2) [right=8em of Z1] {$z_{t+1}$};
\node[stochastic] (Z3) [right=8em of Z2] {$z_{t+2}$};

\node[deterministic] (H1) [below=8em of Z1] {$h_t$};
\node[deterministic] (H2) [below=8em of Z2] {$h_{t+1}$};
\node[deterministic] (H3) [below=8em of Z3] {$h_{t+2}$};

\node[invisible] (past) [left=5em of H1] {$\dots$};
\node[invisible] (future) [right=5em of H3] {$\dots$};

\node[stochastic] (X1) [below=8em of H1] {$x_t$};
\node[stochastic] (X2) [below=8em of H2] {$x_{t+1}$};
\node[stochastic] (X3) [below=8em of H3] {$x_{t+2}$};

\path[line width=.1em] (Z1) edge (H1);
\path[line width=.1em] (Z2) edge (H2);
\path[line width=.1em] (Z3) edge (H3);

\path[line width=.1em] (X1) edge (H1);
\path[line width=.1em] (X2) edge (H2);
\path[line width=.1em] (X3) edge (H3);

\path[line width=.1em] (H1) edge (X2);
\path[line width=.1em] (H2) edge (X3);

\path[line width=.1em] (past) edge (H1);
\path[line width=.1em] (H1) edge (H2);
\path[line width=.1em] (H2) edge (H3);
\path[line width=.1em] (H3) edge (future);
\end{tikzpicture}
\caption{
    Graphical model corresponding to the factorisation given in Eq.~(\ref{eq:storn_ll}). 
    The hidden states $h_t$ are shown as diamonds to stress that they are no source of stochasticity. 
    Despite of this, marginalising out $\bz\lT$ makes $\bh\lT$ stochastic.}
\end{figure}

The only quantities bearing uncertainty in the calculation of $\bh\lT$ are the latent variables $\bz\lT$, as $\bx\lT$ stems from the data set and for all $t$, $h_t$ is a deterministic function of $\bx\ltt$ and $\bz\ltt$.
The resulting factorisation of the data likelihood of a single sequence $p(\bx\lT)$ is then 
\eq{
    p(\bx\lT)
        &= \prod_{t=0}^{T-1} p(x_{t+1}|\bx\ltt) \\
        &= \int_{\bz\lT} p(\bz\lT) \prod_{t=0}^{T-1} p(x_{t+1}|\bx\ltt, \bz\ltt, \cancel{\bz_{t+1:T}}) d\bz\lT\\
        &= \int_{\bz\lT} p(\bz\lT) \prod_{t=0}^{T-1} \int_{h_{t}} p(x_{t+1}|\bx\ltt, \bz\ltt, h_{t}) p(h_{t}|\bx\ltt, \bz\ltt) d h_t d\bz\lT,
}
where we have made use of the fact that $x_{t+1}$ is independent of $\bz_{t+1:T}$.
Since $h_t$ is a deterministic function of $\bx\ltt$ and $\bz\ltt$, we note that $p(h_t|\bx\ltt, \bz\ltt)$ follows a Dirac distribution with its mode given by Eq.~(\ref{eq:hidden_storn}).
Thus, the integral over the hidden states is replaced by a single point; we make the dependency of $h_t$ on both $\bz\ltt$ and $\bx\ltt$ explicit.
\eq{
    p(\bx\lT)
        &= \int_{\bz\lT} p(\bz\lT) \prod_{t=0}^{T-1} p(x_{t+1}|h_t(\bx\ltt, \bz\ltt)) d\bz\lT. \numberthis \label{eq:storn_ll}
}
The corresponding graphical model is shown in Figure~\ref{fig:graphmodel}.
Even though the determinism of $h_t$ might seem restrictive at first, we will argue that it is not.
Let $\bh\lT$ be the sequence of hidden layer activations as given by Eq.~(\ref{eq:hidden_storn}).
This sequence is deterministic given $\bx\lT$ and $\bz\lT$ and consequently, $p(\bh\lT|\bx\lT, \bz\lT)$ will follow a Dirac distribution. 
Marginalising out $\bz\lT$ will however lead to a universal approximator of probability distributions over sequences, analoguously to the argument given in Section~\ref{sec:sgvb}.

An additional consequence is, that we can restrict ourselves to prior distributions over the latent variables that factorise over time steps, i.e. $p(\bz\lT) = \prod_t p(z_t)$.
This is much easier to handle in practice, as calculating necessary quantities such as the KL-divergence can be done independently over all time steps and components of $z_t$.

Despite of this, the distribution over $\bh\lT$ will be a Markov chain and can exhibit stochastic behaviour, if necessary for modelling the data distribution.

\subsection{Variational Inference for Latent State Sequences}
The derivation of the training criterion is done by obtaining a variational upper bound on the negative log-likelihood via Jensen's inequality, where we use a variational approximation $q(\bz\lT|\bx\lT) \approx p(\bz\lT|\bx\lT)$.
\eq{
- \log p(\bx\lT) 
        &= -\log \int_{\bz\lT} {q(\bz\lT|\bx\lT) \over q(\bz\lT|\bx\lT)} p(\bz\lT) \prod_{t=0}^{T-1} p(x_{t+1}|h_t(\bx\ltt, \bz\ltt)) d\bz\lT \\
    &\le KL(q(\bz\lT|\bx\lT)|p(\bz\lT)) - \Expc{\bz\lT \sim q(\bz\lT|\bx\lT)}{\sum_{t=0}^{T-1} \log p(x_t|h_{t-1}, \bz_{1:t})} \numberthis \label{eq:srn-bound} \\
    &:= \mathcal{L}_\text{STORN}
}
In this work, we restrict ourselves to a standard Normal prior\footnote{In a preliminary report, we proposed the use of a Wiener process for a prior. However, the presented results were invalid due to implementation errors and the paper has been withdrawn.} of the form
\eq{
p(\bz\lT) = \prod_{t, k} \mathcal{N}(z_{t, k}|0, 1),
}
where $z_{t, k}$ is the value of the $k$'th latent sequence at time step $t$.

The recognition model $q$ will in this case be parameterised by a single mean $\mu_{t, k}$ and variance $\sigma^2_{t, k}$ for each time step and latent sequence.
Both will be represented by the output of a recurrent net, which thus has $2\omega$ outputs of which the first $\omega$ (representing the mean) will be unconstrained, while the second $\omega$ (representing the variance) need to be strictly positive.
Given the output $\by\lT = f^r(\bx\lT)$ of the recognition \ac{RNN} $f^r$, we set 
\eq{
    \mu_{t, k} &= y_{t, k}, \\
    \sigma^2_{t, k} &= y_{t, k+\omega}^2.
}
Note that the square ensures positiveness.

\begin{figure}
\label{fig:compmodel}
\centering
\tiny
\begin{tikzpicture}[
    -latex,
    auto,
    node distance=12em,
    on grid,
    semithick]

\tikzstyle{data} = [rectangle, draw=black, fill=magenta, text=black, minimum width=3em, minimum height=3em, line width=.1em]

\tikzstyle{sample} = [rectangle, draw=black, fill=green, text=black, minimum width=3em, minimum height=3em, line width=.1em]

\tikzstyle{invisible} = [draw=none, fill=none, text=black]

\tikzstyle{gen} = [rectangle, draw=black, fill=teal, text=white, minimum width=3em, minimum height=3em, line width=.1em]

\tikzstyle{recog} = [rectangle, draw=black, fill=cyan, text=white, minimum width=3em, minimum height=3em, line width=.1em]

\node[data] (X1) {$x_t$};
\node[data] (X2) [right=10em of X1] {$x_{t+1}$};
\node[data] (X3) [right=10em of X2] {$x_{t+2}$};

\node[recog] (HR1) [below=5em of X1] {$h_t^r$};
\node[recog] (HR2) [below=5em of X2] {$h_{t+1}^r$};
\node[recog] (HR3) [below=5em of X3] {$h_{t+2}^r$};

\node[recog] (Q1) [below=5em of HR1] {$y_t^g$};
\node[recog] (Q2) [below=5em of HR2] {$y_{t+1}^g$};
\node[recog] (Q3) [below=5em of HR3] {$y_{t+2}^g$};

\node[sample] (Z1) [below=5em of Q1] {$z_t$};
\node[sample] (Z2) [below=5em of Q2] {$z_{t+1}$};
\node[sample] (Z3) [below=5em of Q3] {$z_{t+2}$};

\node[gen] (HG1) [below=5em of Z1] {$h_t^g$};
\node[gen] (HG2) [below=5em of Z2] {$h_{t+1}^g$};
\node[gen] (HG3) [below=5em of Z3] {$h_{t+2}^g$};

\node[gen] (Y1) [below=5em of HG1] {$y_t^g$};
\node[gen] (Y2) [below=5em of HG2] {$y_{t+1}^g$};
\node[gen] (Y3) [below=5em of HG3] {$y_{t+2}^g$};

\node[invisible] (pastr) [left=5em of HR1] {$\dots$};
\node[invisible] (futurer) [right=5em of HR3] {$\dots$};

\node[invisible] (pastg) [left=5em of HG1] {$\dots$};
\node[invisible] (futureg) [right=5em of HG3] {$\dots$};

\path[line width=.1em] (X1) edge (HR1);
\path[line width=.1em] (X2) edge (HR2);
\path[line width=.1em] (X3) edge (HR3);

\path[line width=.1em] (HR1) edge (Q1);
\path[line width=.1em] (HR2) edge (Q2);
\path[line width=.1em] (HR3) edge (Q3);

\path[line width=.1em] (Q1) edge (Z1);
\path[line width=.1em] (Q2) edge (Z2);
\path[line width=.1em] (Q3) edge (Z3);

\path[line width=.1em] (Z1) edge (HG1);
\path[line width=.1em] (Z2) edge (HG2);
\path[line width=.1em] (Z3) edge (HG3);

\draw (X1) edge[line width=.1em, out=330,in=135] (HG2);
\draw (X2) edge[line width=.1em, out=330,in=135] (HG3);
\draw (X3) edge[line width=.1em, out=330,in=135] (futureg);

\path[line width=.1em] (HG1) edge (Y1);
\path[line width=.1em] (HG2) edge (Y2);
\path[line width=.1em] (HG3) edge (Y3);

\path[line width=.1em] (pastr) edge (HR1);
\path[line width=.1em] (HR1) edge (HR2);
\path[line width=.1em] (HR2) edge (HR3);
\path[line width=.1em] (HR3) edge (futurer);

\path[line width=.1em] (pastg) edge (HG1);
\path[line width=.1em] (HG1) edge (HG2);
\path[line width=.1em] (HG2) edge (HG3);
\path[line width=.1em] (HG3) edge (futureg);

\draw (Y1) edge[line width=.1em, color=red,out=60,in=300] (X1);
\draw (Y2) edge[line width=.1em, color=red,out=60,in=300] (X2);
\draw (Y3) edge[line width=.1em, color=red,out=60,in=300] (X3);

\end{tikzpicture}
\caption{
    Diagram of the \emph{computational} dependencies of \acp{STORN}.
    Each node of the graph corresponds to a vectorial quantity.
    The different types of nodes shown are data (\textcolor{magenta}{magenta}), the recognition model (\textcolor{cyan}{cyan}), samples (\textcolor{green}{green}) and the generating model (\textcolor{teal}{teal}).
   Note that the outputs of the recognition model $y^r_t$ depict the statistics of $q(z_t|\bx\ltt)$, from which the sample $z_t$ (green) is drawn.
    The output of the generating model, $y^g_t$ is used to represent $p(x_{t+1}|\bx\ltt)$.
    The red arrow expresses that this prediction is used to evaluate the loss, i.e. the negative log-likelihood.
}
\end{figure}
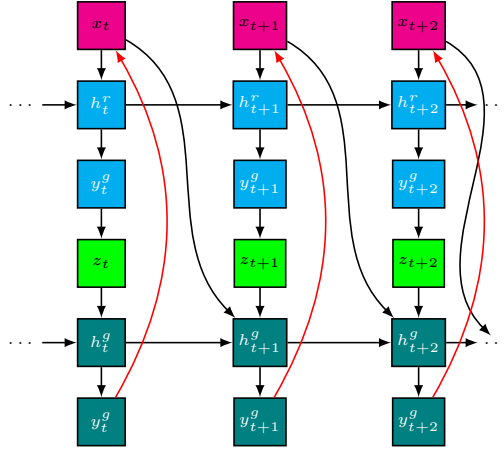

Going along with the reparametrisation trick of \cite{kingma2013auto}, we will sample from a standard Normal at each time step, i.e. $\epsilon_{t, k} \sim \mathcal{N}(0, 1)$ and use it to sample from $q$ via $z_{t, k} = \mu_{t, k} + \sigma_{t, k}\epsilon_{t, k}$.
Given the complete sample sequence $\bz\lT$ we calculate the two terms of Equation~(\ref{eq:srn-bound}). 
The KL-divergence can be readily computed, while we need to pass $\bz\lT$ through the  generating model $f^g$ which gives $-\log p(\bx\lT|\bz\lT)$.
The computational flow is illustrated in Figure~\ref{fig:compmodel}.

\subsection{Comparison to \acp{RNN}} An important question is whether the proposed model offers any theoretical improvements over \acp{RNN} with no latent variables.
The approximation capabilities (with respect to probability distributions) of \acp{RNN} result from the choice of likelihood function, i.e. the way the density of the observations at time step $t$ is determined by the outputs of the network, $y_t$. See Eq.~(\ref{eq:bernoulli}).
We have argued in Section~\ref{sec:intro} that a na\"ive Bayes assumption reduces the approximation capabilities.
One way to circumvent this is to use mixture distributions 
\citep{graves2013generating}.
The number of parameters of the latter scales poorly, though: linear in the number of modes, hidden units in the last layer and output dimensions.

Both approaches also share the drawback that the stochasticity entering the computation is not represented in the hidden layers: drawing a sample is determined by a random process invisible to the network.

\ac{STORN} overcomes both of these issues.
Introducing an additional mode merely requires an additional change of curvature in the approximation of $F$ (compare Section~\ref{sec:sgvb}).
This can be obtained by additional hidden units, for which the number of parameters scales linearly in the number of hidden units in the incoming and outgoing layer.
Further, the stochasticity in the network is stemming from $z$ of which the hidden layer is a function.

\section{Experiments}
For evaluation we trained the proposed model on a set of midi music, which was used previously \citep{bengio2012advances,pascanu2013construct,bayer2013fast,boulanger2012modeling} to evaluate \acp{RNN}.
We also investigated modelling human motion in the form of motion capture data  \citep{boulanger2012modeling,sutskever2008recurrent,taylor2006modeling}.
We employ \acp{FD-RNN}~\citep{bayer2013fast} for both the recognition and the generating model.
While we determine the dropout rates for the generating model via model selection on a validation set, we include them into the parameter set for the recognition model.
In a manner similar to \cite{bayer2013training}, we exploit fast dropout's natural inclusion of variance as the variance for the recognition model, i.e. $\sigma^2_{t, k}$.
We used Adadelta~\citep{zeiler2012adadelta} enhanced with Nesterov momentum~\citep{sutskever2013importance} for optimisation.

\subsection{Polyphonic Music Generation}

All experiments were done by performing a random search~\citep{bergstra2012random} over the hyper parameters, where 128 runs were performed for each data set. 
Both the recognition and the generating model used 300 hidden units with the logistic sigmoid as the transfer function.
We report the estimated negative log-likelihood (obtained via the optimiser proposed in \citep{rezende2014stochastic}) on the test set of the parameters which yielded the best bound on the validation set.

As expected, \ac{STORN} improves over the models assuming a factorised output distribution (FD-RNN, sRNN, Deep RNN) in all cases.
Still, RNN-NADE has a competitive edge.
The reasons for this remain unclear from the results alone, but the stochastic training and resulting noisy gradients are a viable hypothesis, since RNN-NADE does not suffer from those.

\begin{table}
    \caption[ ]{
        Results on the midi data sets. 
        All numbers are average negative log-likelihoods on the test set, where ``FD-RNN'' represents the work from \cite{bayer2013fast}; ``sRNN'' and ``RNN-NADE'' results are from \cite{bengio2012advances} while ``Deep RNN`` shows the best results from \cite{pascanu2013construct}. 
    The results of our work are shown as ``STORN`` and have been obtained by means of the importance sampler described in~\cite{rezende2014stochastic}.
    }
    \label{table:results-midi}
    \begin{center}
    \begin{small}
    \begin{tabular}[t]{|l|r|r|r||r|r|}
        \hline
        Data set & STORN & FD-RNN & sRNN & RNN-NADE & Deep RNN \\
        \hline
        Piano-midi.de & 7.13 & 7.39 & 7.58 & 7.05 & -- \\
        Nottingham & 2.85 & 3.09 & 3.43 & 2.31 & 2.95  \\
        MuseData & 6.16 & 6.75 & 6.99 & 5.60 & 6.59 \\
        \belowspace
        JSBChorales & 6.91 & 8.01 & 8.58 & 5.19 & 7.92 \\
        \hline
    \end{tabular}
    \end{small}
    \end{center}
\end{table}

\subsection{Motion Capture Data}

The motion capture data set \citep{hsu2005style,taylor2006modeling}  is a sequence of kinematic quantities obtained from a human body during walking.
It consists of 3128 time steps of 49 angular quantities each.
The experimental protocol of previous studies of this data set is to report the mean squared error on the training set , which we comply with.
\footnote{The use of the MSE on the trainig set is debatable for this task. First, there is the danger of overfitting the training set. Second, the metric only captures a single moment of the residual distribution. We go forward with this protocol nonetheless to make our results comparable to previous works. Additionally, we report the negative log-likelihood, which is the right metric for the task.}

For motion capture data, we chose a Gaussian likelihood with a fixed standard deviation for the generating model.
The recognition model was chosen to be a bidirectional RNN.
While the standard deviation was fixed to $1$ during training, we performed a binary search for a better value after training; the resulting estimate of the negative log-likelihood on the validation set was then used for model selection.

\begin{figure}
\begin{center}
    \includegraphics[width=\textwidth]{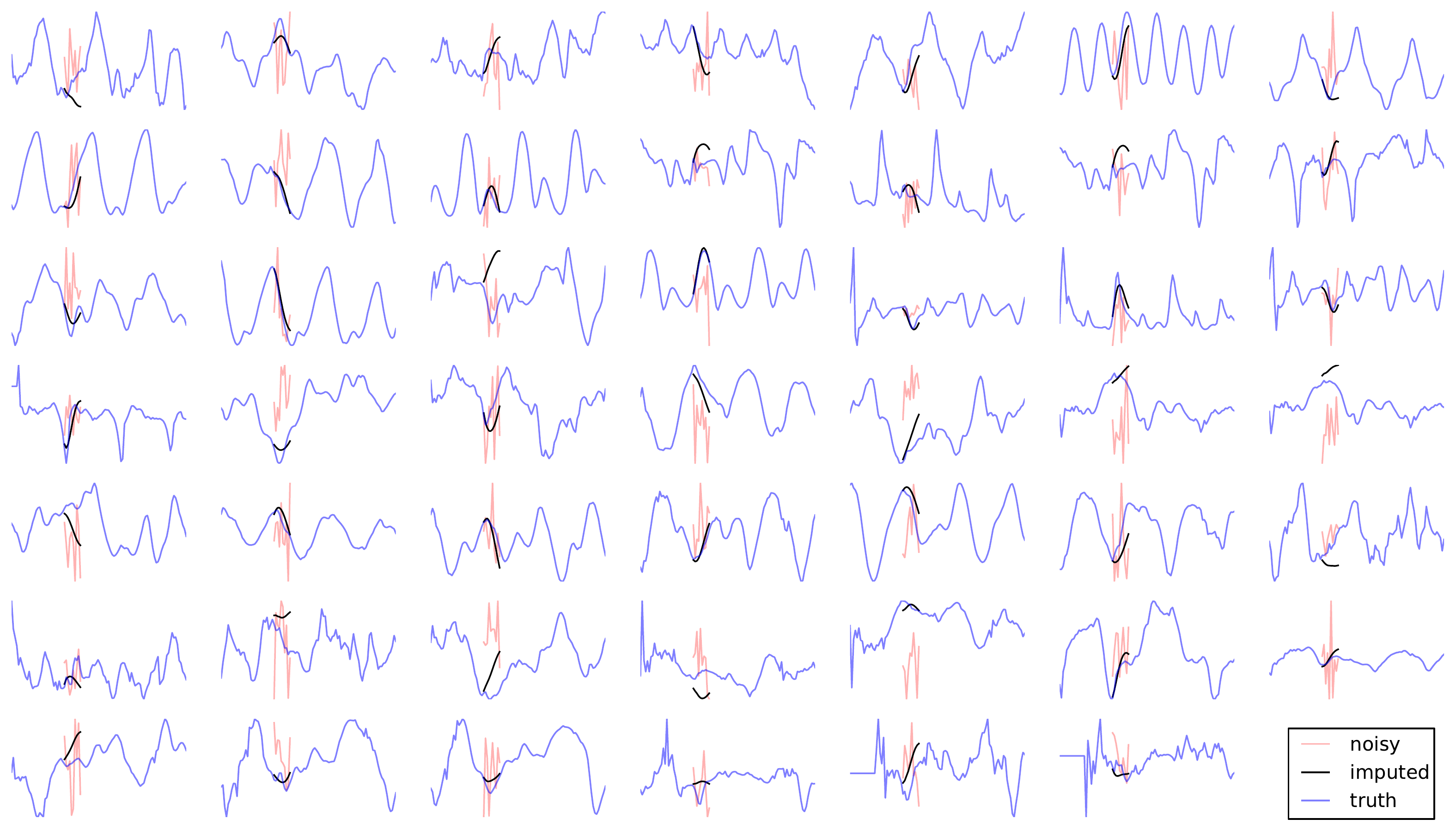} \\
\end{center}
\caption{
    Illustration of missing value imputation on the motion capture data set.
    We show the first 48 of the 49 channels of a random sample, where time steps 30 to 40 were initialised with random noise.
    Subsequently, a maximum a posteriori point estimate of the latent variables was used to reconstruct the missing parts of the signals.
}
\label{fig:impute}
\end{figure}

The estimated negative log-likelihood of the data was 15.99.
Other models trained on this data set, namely the RNN-RBM, RTRBM and cRBM do not offer a tractable way of estimating the log-likelihood of the data, which is why there is no direct mean of comparison respecting the probabilistic nature of the models.
In the case of the former two, the mean squared prediction error is reported instead, which is 20.1 and 16.2 respectively.
Our method achieved an average MSE of 4.94, which is substantially less than previously reported results.For additional means of comparison, we performed approximate missing value imputation of motion capture data.
We picked random sequences of length 60 and replaced all of the 49 channels from time steps 30 to 40 with standard normal noise.
We then performed a \emph{maximum a posteriori} point selection of the recognition model, i.e. $\argmax_{\hat{\bz}\lT} q(\hat{\bz}\lT|\bx\lT)$, from which we reconstructed the output via $\argmax_{\hat{\bx}_{30:40}} \log p(\bx\lT|\hat{\bz}\lT)$.  
Note that this method is different from the one proposed in \citep{rezende2014stochastic}, where an iterative scheme is used.
We also experimented with that method, but did not find it to yield substantially better results.
The results of the imputations are shown in Figure~\ref{fig:impute}.

\begin{figure}
\begin{center}
    \includegraphics[width=\textwidth]{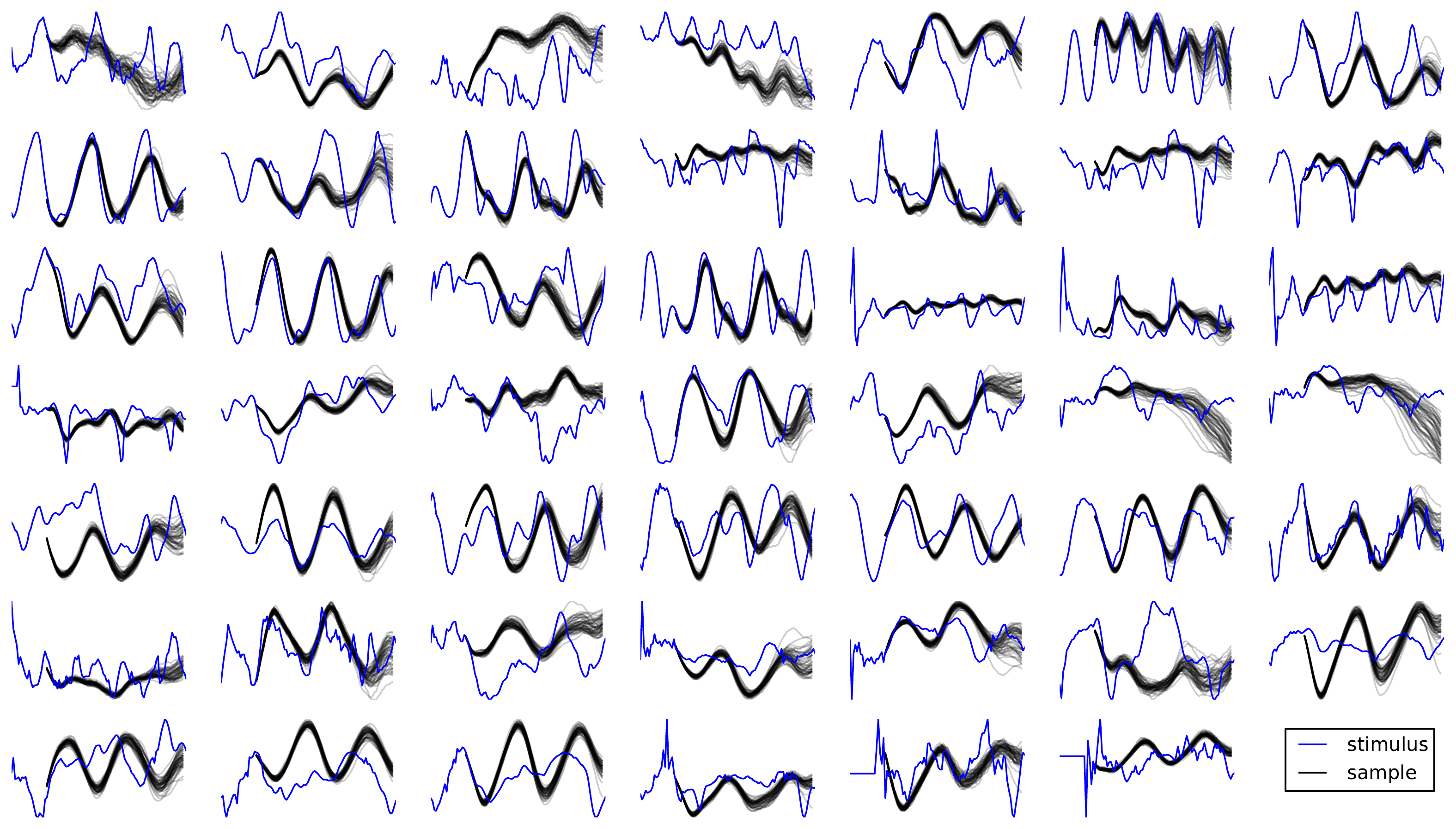} \\
\end{center}
\caption{
    Samples from the model trained on motion capture data after providing a stimulus prefix sequence of 20 time steps.
    The uncertainty of the learned distribution is visible by the diversity of the samples; nevertheless, the distribution is rather unimodal.
}
\label{fig:sample}
\end{figure}

To demonstrate the generative capabilities of the method, we drew 50 samples from the model after initialising it with a stimulus prefix.
The stimulus had a length of 20, after which we ran the model in ``generating mode'' for another 80 time steps.
This was done by feeding the mean of the model's output at time step $t$ into the generating model at time step $t+1$. 
Additionally, we drew $\bz_{20:80}$ from the prior. The results are visualised in Figure~\ref{fig:sample}.

\section{Discussion and Future Work}
We have presented a model class of stochastic \acp{RNN} that can be trained with a recently proposed estimator, \ac{SGVB}.
The resulting model fulfills the expectation to greatly improve over the performance of \acp{sRNN} erroneously assuming a factorisation of output variables.
An important take away message of this work is that the performance of \acp{RNN} can greatly benefit from more sophisticated methods that greatly improve the representative capabilities of the model.

While not shown in this work, \acp{STORN} can be readily extended to feature computationally more powerful architectures such as LSTM or deep transition operators \citep{hochreiter1997long, pascanu2013construct}.

Still, an apparent weakness seems to be the stochastic objective function.
Thankfully, research in optimisation of stochastic objective functions has far from halted and we believe \ac{STORN} to benefit from any advances in that area.

\section{Acknowledgements}
Part of this work has been supported by the TACMAN project, EC Grant agreement no. 610967, within the FP7 framework programme. 

\begin{acronym}[YTB]
    \acro{NN}{Neural Network}
    \acro{RNN}{Recurrent Neural Network}
    \acro{sRNN}{simple Recurrent Neural Network}
    \acro{LSTM}{Long short-term memory}
    \acro{MSE}{mean squared error}
    \acro{FFN}{feed forward neural network}
    \acro{SGVB}{stochastic gradient variational Bayes}
    \acro{FD-RNN}{Fast Dropout Recurrent Network}
    \acro{STORN}{Stochastic Recurrent Network}
\end{acronym}

\bibliographystyle{iclr2015}
\bibliography{tex/bibliography}{}

\end{document}